%% file: PaperForReview.tex
\documentclass[10pt,twocolumn,letterpaper]{article}

\usepackage{wacv}              

\usepackage{graphicx}
\usepackage{amsmath}
\usepackage{amssymb}
\usepackage{booktabs}

\usepackage{symmath}
\usepackage{multirow}
\usepackage{makecell}
\usepackage{pifont}
\usepackage[dvipsnames]{xcolor}
\usepackage{pythonhighlight}
\usepackage[accsupp]{axessibility} 

\usepackage{color, colortbl} 

\usepackage{tabu}           
\usepackage{adjustbox}
\usepackage{bm}

\input{math_commands}

\definecolor{mydarkblue}{rgb}{0,0.53,0.96}
\newcommand{\layer}[1]{\ensuremath{\mathsf{#1}\xspace}}
\newcommand{\subsec}[1]{\noindent\textbf{#1}~~}

\newcommand{\cmark}{\textcolor{OliveGreen}{\ding{51}}}
\newcommand{\xmark}{\textcolor{BrickRed}{\ding{55}}}

\newcommand{\increasenoparent}[1]{\textcolor{ForestGreen}{+#1}}

\usepackage[pagebackref,breaklinks,colorlinks]{hyperref}
\usepackage[capitalize]{cleveref}
\crefname{section}{Sec.}{Secs.}
\Crefname{section}{Section}{Sections}
\Crefname{table}{Table}{Tables}
\crefname{table}{Tab.}{Tabs.}
\usepackage[accsupp]{axessibility}  

\def\wacvPaperID{1223} 
\def\confName{WACV}
\def\confYear{2024}

\newcommand{\papertitle}{Fast and Interpretable Face Identification for Out-Of-Distribution Data Using Vision Transformers}

\begin{document}

\title{\papertitle}


\author{Hai Phan$^1$ \hspace{0.5cm} Cindy X. Le$^2$ 
\hspace{0.5cm} Vu Le$^3$ \hspace{0.5cm} Yihui He$^4$ \hspace{0.5cm} Anh ``Totti'' Nguyen$^1$\\
{\tt\small \{pthai1204, anh.ng8, leducvuvietnam\}@gmail.com \hspace{0.5cm} xl2738@columbia.edu \hspace{0.5cm} 
he2@alumni.cmu.edu}\\
\hspace{1.0cm}$^1$Auburn University \hspace{0.5cm} $^2$Columbia University \hspace{0.5cm} $^3$Phenikaa University  \hspace{0.5cm} $^4$Carnegie Mellon University
}
\maketitle

\begin{abstract}
Most face identification approaches employ a Siamese neural network to compare two images at the image embedding level.
Yet, this technique can be subject to occlusion (\eg, faces with masks or sunglasses) and out-of-distribution data.
DeepFace-EMD \cite{hai2022deepface} reaches state-of-the-art accuracy on out-of-distribution data by first comparing two images at the image level, and then at the patch level.
Yet, its later patch-wise re-ranking stage admits a large $O(n^3 \log n)$ time complexity (for $n$ patches in an image) due to the optimal transport optimization.

In this paper, we propose a novel, 2-image Vision Transformers (ViTs) that compares two images at the patch level using cross attention.
After training on 2M pairs of images on CASIA Webface \cite{yi2014learning}, our model performs at a comparable accuracy as DeepFace-EMD on out-of-distribution data, yet at an inference speed more than twice as fast as DeepFace-EMD \cite{hai2022deepface}. 
In addition, via a human study, our model shows promising explainability through the visualization of cross-attention. 
We believe our work can inspire more explorations in using ViTs for face identification.
\end{abstract}

\section{Introduction}
\label{sec:intro}

Face identification (FI), the technology that enables automatic identification of individuals from photographs, is widely used in law enforcement \cite{MiaSato.2021, airplane, lawsuit2021facial}, private businesses \cite{barredgrocerystore}, smartphones\cite{applefaceid}, and so on. With growing data volumes, fast and high FI systems are paramount for processing and analyzing real-time data to identify faces and patterns effectively. 

Unfortunately, facial information may not always be obtained in ideal conditions, and out-of-distribution data (OOD) \eg faces with masks, sunglasses, or other adversarial components, poses challenges to correctly identifying the targets. FI accuracy may drop substantially on OOD data, \eg, from 98.41\% to 39.79\% on LFW when the query face is wearing masks \cite{hai2022deepface} or adversarially modified \cite{zhong2020towards,amos2016openface}. 

Besides the accuracy of FI for OOD data, the field faces two practical challenges. The first challenge is the rapid identification of faces under OOD settings. Swift identification can improve user experience by reducing waiting time during unlocking devices, accessing accounts \cite{applefaceid}, and security checks \cite{barredgrocerystore}, increasing people's trust towards machine-generated results \cite{efendic2020slow}, and lowering emergency response \cite{filimiations}.
The second challenge is how to explain FI decisions to the end-users, which is interestingly understudied. In reality, FI systems are often operated by end-users \cite{phillips2018face} who expect to get real-time answers and the reasons why such answers are given. The current limited machine-user interoperability causes numerous false decisions \cite{lawsuit2021facial,newjersey2021arrested,detroit2021arrested,michigan2021arrested}.
Specifically, only a few studies have produced explanations for FI predictions \cite{hai2022deepface,stylianou2019visualizing} and none have evaluated the explanations from interpretable FI models on users.

\begin{figure}[!t]
    \centering
    \includegraphics[width=0.5\textwidth]{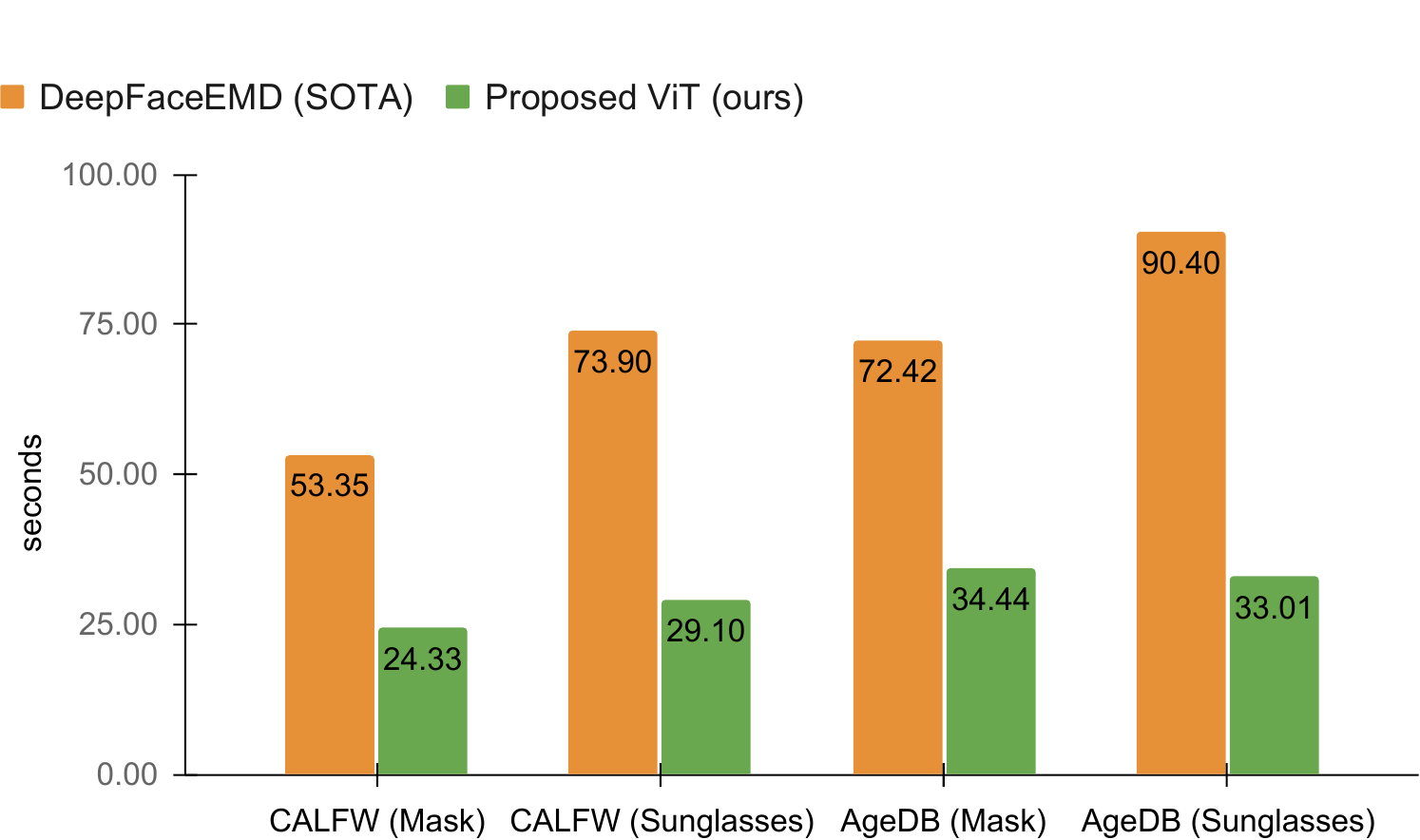}
    \caption{Actual running time in seconds (lower is better) for the re-ranking computation in face identification under occlusion. 
    Our proposed model is at least two times faster than the state-of-the-art  DeepFace-EMD \cite{hai2022deepface} over all the datasets.}
    \label{fig:model_inference_time_on_ood}
\end{figure}

\begin{figure*}
    \centering
    \includegraphics[width=1.0\textwidth]{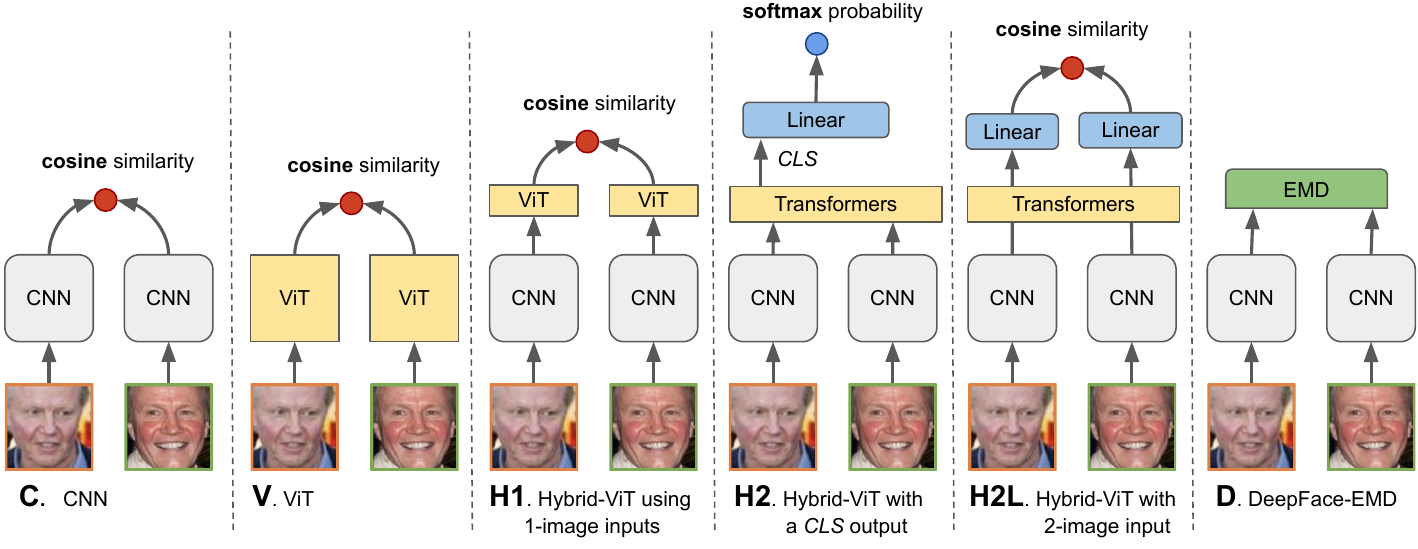}
    \caption{
    The architecture of the six networks evaluated in this work including our proposed H2L. 
    } 
    \label{fig:approaches}
\end{figure*}

In this work, we explore the design space of ViTs that enable cross-image attention between two input images for FI.
On three important criteria (1) accuracy on in-distribution and OOD data, (2) computational complexity, and (3) explainability, we compare ViTs, CNNs, and EMD-based patch-wise re-ranking methods (\cref{fig:approaches}) and find that: 
\footnote{Code, demo and data are available at \url{https://github.com/anguyen8/face-vit}}
\begin{itemize}
    \item With cross-image attention, our 2-image Hybrid-ViT model is an effective re-ranking approach. It outperforms traditional FI models (based on CNNs and 1-image ViTs) on both in-distribution and OOD data. 
    \item Our 2-image Hybrid-ViT performs on par with DeepFace-EMD \cite{hai2022deepface}---a state-of-the-art approach to OOD face identification. 
    In addition, our proposed model is more scalable as shown in \cref{fig:model_inference_time_on_ood}, \ie running over 2$\times$ faster in practice than DeepFace-EMD, which is slow due to the optimal transport optimization phase (\cref{sec:time_complextity}).
    
    \item In a 21-person human study, the users of Hybrid-ViTs and DeepFace-EMD explanations scored substantially higher than the users of Siamese neural networks (SNNs) in face verification (\cref{sec:explainability}).
    We are the first to report that visual explanations improve end-user accuracy in face verification.
\end{itemize}

To our knowledge, our work is the first to (1) explore the design space of ViTs \cite{dosovitskiy2020vit} for the FI problems on OOD data; (2) compare ViT-based and EMD-based image similarity approaches \cite{hai2022deepface,Zhang_2020_CVPR,zhang2020deepemd,zhao2021towards}; and (3) study how visual explanations improve human accuracy in face verification.

\section{Related Work}
\label{sec:related_work}

For face recognition, previous deep approaches typically adopt a CNN architecture (\eg VGGNet \cite{Simonyan-VGGNet}, ResNet \cite{Kaiming_ResNet}, etc.) as the backbone to extract deep face features and then use metric learning methods \cite{schroff2015facenet, deng2018arcface} to classify identities. This design often achieves impressive results for in-distribution but fails on OOD data. A recent work, DeepFace-EMD \cite{hai2022deepface}, provided an Earth Mover's Distance (EMD) distance to obtain cross-image information from CNN outputs, improving face OOD. Similar to DeepFace-EMD, we explore Transformers to exploit cross-attention information between inputs.

\subsec{Out-of-distribution face identification.} 
Identifying faces under occlusion \cite{hai2022deepface,wang2021mlfw,qiu2021end2end} or adversarial changes \cite{zhong2020towards} is challenging. FI systems using SNNs are vulnerable to images containing sunglasses, masks, or adversarial perturbations. 
A line of approach re-trains deep CNN feature extractors on images with partially-occluded faces \cite{trigueros2018enhancing,wang2021mlfw,osherov2017increasing,xu2020improving,guo2018face,xu2020improving}.
However, data augmentation on a specific type of occlusion (\eg face masks) does not guarantee generalization to new OOD changes (\eg in hairstyles) in the input image \cite{hai2022deepface}.
An alternative technique for OOD face data is to reconstruct the missing pixels before performing FI \cite{wright2008robust,zhou2009face,yang2011robust,he2011regularized,li2013structured,Zhao2018RobustLF}.
Yet, the de-occlusion process \cite{dong2020occlusion,cai2020semi,ge2020occluded} may fail to preserve the identity of the target person and add another level of abstraction over how the FI system computes its decisions, further opaquing the decision-making process.

\subsec{Siamese networks for patch-wise comparison.}
A common FI technique involves adopting the Siamese architecture, feeding a pair of input images into two weight-shared, CNN-based feature extractors, and comparing the cosine similarity between two output image-level embeddings  \cite{Liu_2017_CVPR,deng2018arcface,schroff2015facenet,wang2018cosface}. Recent EMD-based image similarity work found that combining both image-level and patch-level similarity yields higher accuracy on in-distribution data \cite{zhao2021towards} and OOD data \cite{zhang2020deepemd,hai2022deepface}. DeepFace-EMD \cite{hai2022deepface} consistently outperforms traditional methods \cite{wang2018cosface,deng2018arcface,schroff2015facenet} that are based on the cosine similarity of two image embeddings from a SNN. Such approaches, however, only conduct a global, image-level comparison and may discard useful local, patch-level information. Researchers are looking for more accurate and efficient architectures for FI tasks.

\subsec{Vision Transformers for patch-wise comparison.} Operating at the patch level, 
ViTs are increasingly popular in computer vision \cite{dosovitskiy2020vit,koner2021oodformer,pmlr-v139-touvron21a,zhu2020deformable}, were shown to achieve remarkable image classification accuracy, and do not need explicit feature extraction like in CNN-based models. Most ViT research focuses on a \emph{single}-image architecture where self-attention \cite{vaswani2017attention} is leveraged to compare the similarity between \emph{intra-image} patches \cite{zhong2021face} or between image-patches and text-tokens in image-text architectures \cite{kim2021vilt,li2021align}.
CrossViT \cite{chen2021crossvit} proposed to use two Transformers but for two differently scaled versions of \emph{the same} image, not for two images.
The only work utilizing ViTs in FI that we are aware of is the concurrent work by \cite{zhong2021face}, which uses the vanilla ViT on a \emph{single} image and therefore offers no cross-image interaction.
A few other concurrent works also explore ViTs for 2-image inputs but rather for person re-identification \cite{wang2022nformer,liao2021transmatcher}, a different task that involves a more unconstrained image distribution than the images typically cropped and aligned in FI. These leave us great room for exploring cross-image interaction to compare two face images.

\subsec{Model interpretability of Vision Transformers.}
Various efforts have been made to visualize the effects of ViTs. Black et al.~\cite{Black2022SimilarityTransformer} proposed a novel method to combine cross-correlation and an attention flow approximation between two images, each processed by a different 1-image ViT.
For multimodal, vision-language Transformers, Kim et al.~\cite{kim2021vilt} use the similarity flow between text and image tokens as explanations for its similarity score.
Chefer et al.~\cite{chefer2021generic,chefer2021transformer} leveraged the aggregate cross-attention across layers and its gradients to derive a visualization of similarity between two inputs.
In our work, we visualize all ViTs using the technique proposed by \cite{stylianou2019visualizing}.

\section{Method}
\label{sec:methods}



We propose a novel ViT architecture (denoted as Model H2L) for FI on OOD data. It takes in \emph{two} images as input to leverage both self-attention and cross-attention to compute a similarity score for two images. 

\subsection{Problem Formulation}

Similar to DeepFace-EMD \cite{hai2022deepface}, our method identifies a person in a query image by ranking all gallery images based on their pair-wise similarity with the query. After ranking (ST1) or re-ranking (ST2), we take the top-1 nearest image as the predicted identity.
For the scope of this paper, we only consider data consisting of frontal faces without gestures.

\subsection{Architecture: a two-Image Hybrid ViT}

\begin{figure}[!ht]
    
    \centering
    \includegraphics[width=0.5\textwidth]{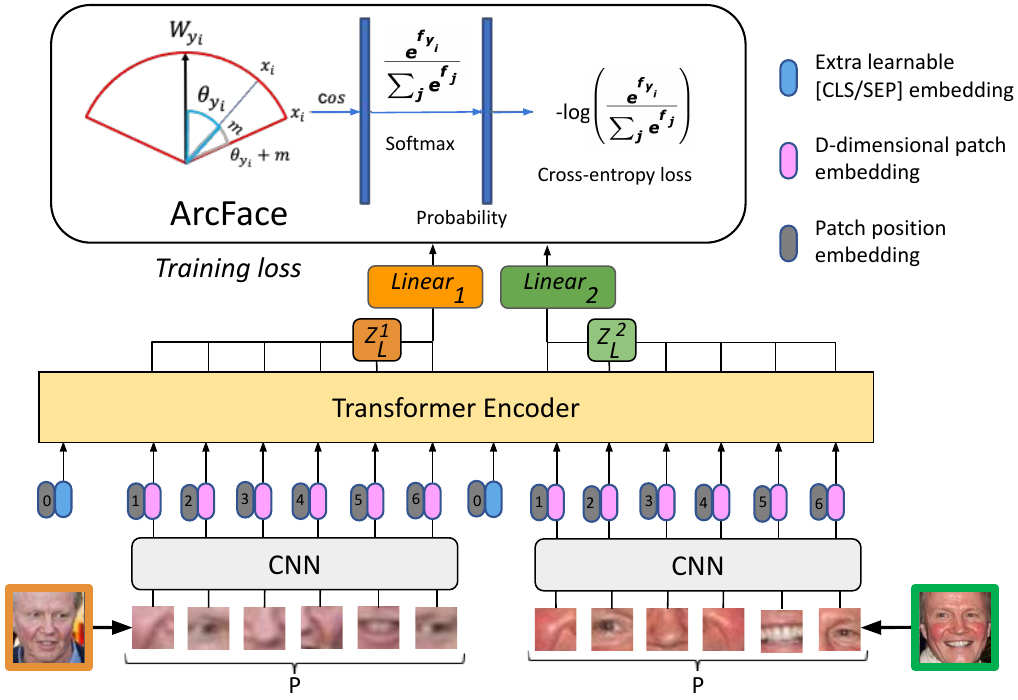}
    \caption{
    The architecture of the proposed ViT-based Model H2L.}
    \label{fig:framework}
\end{figure}

The overall architecture of the model is shown in \cref{tab:networks}. and \cref{fig:framework}. It takes in patch embeddings from a pre-trained CNN (ArcFace \cite{deng2018arcface}). The Transformer encoder consists of a block of a multiheaded self-attention (MSA) layer and an MLP layer. After $N$ layers of the Transformer encoder, which contains both self-attention and cross-attention from 2 input images, the patch embeddings of the input images go through two linear layers.

\subsec{Face embeddings.}
For a zero-shot face problem, deep metric learning works efficiently \cite{schroff2015facenet,Liu_2017_CVPR,wang2018cosface}. Besides \textit{[CLS]} (classification tokens) \cite{devlin2018bert,dosovitskiy2020vit} for feature embeddings, we also use the remaining 2-output to separate linear layers to extract features that are deployed to a deep metric learning fashion (see \cref{fig:framework} for details). 

Given two input 2D face images $\mathbf{x}_1, \mathbf{x}_2$, we reshape them to have dimensions  $\in \mathbb{R}^{H\times W \times C}$. The face embeddings $\mathbf{x}_{p1}, \mathbf{x}_{p2} \in \mathbb{R}^{P^2 \times D}$ are extracted from either CNNs or a linear embedding layer, where $P$ is the number of the face patches and $D$ is the size of the patch embedding. Here in the loss function ArcFace \cite{deng2018arcface}, we use $D=512$ and $P=8$.  

We denote the learnable embeddings as $\textbf{E} \ \text{and} \ \textbf{E}_{pos} \in \mathbb{R}^{(2\times P^2 + 2)  \times D}$, the two extra learnable embeddings as $\textbf{X}_{CLS}$ and $\textbf{X}_{SEP}$, and the intermediate layers of the Transformer encoder as $\textbf{z}_{i}$. $\textbf{f}_1$ and $\textbf{f}_2$ are the features from \textit{two} linear layers that contain cross-attention information between two images. Our proposed two-image-based model can be formulated as follows. 

\begin{align}
\label{eq:transfromer}
    \textbf{z}_0 &= [\mathbf{x}_{CLS}\textbf{E},  \mathbf{x}_{p1} \textbf{E}, \mathbf{x}_{SEP} \textbf{E}, \mathbf{x}_{p2} \textbf{E}] + \textbf{E}_{pos},\\
    \textbf{z}^{'}_l &=  \text{MSA}(\text{LayerNorm}(\textbf{z}_{l-1})), \ \ \ l = 1 \dots L \\
    \textbf{z}_l &= \text{MLP}(\text{LayerNorm}(\textbf{z}_{l}^{'})) + \textbf{z}_l^{'}, \ l = 1 \dots L \\
    \textbf{z}_l &\equiv [\textbf{z}_{CLS}, \textbf{z}_L^1, \textbf{z}_{SEP}, \textbf{z}_L^2], \ \ \textbf{z}_L^1, \textbf{z}_L^2 \in \mathbb{R}^{P^2  \times D}\\
    \textbf{f}_1 &= \text{LayerNorm}(\text{Linear}_1(\textbf{z}_L^1))\\
    \textbf{f}_2 &= \text{LayerNorm}(\text{Linear}_2(\textbf{z}_L^2))\\
    \text{loss} &= \text{Arcface\_loss}(\textbf{f}_1, \textbf{f}_2) 
\end{align}

\textbf{Position embeddings} in vanilla Transformers \cite{vaswani2017attention} indicate the position of words in sentences for machine translation. Here, they are also used with the face inputs. When parts of the face are arranged in a constrained order, \eg position of eyes, mouth, etc. this positioning information maintains the facial structure. 

\textbf{Attention-based outputs.} The outputs $\textbf{z}^{'}_l$ from a multi-head-attention (MSA) layer are obtained through a combination of self and cross-attention processes. Previous ViT works \cite{dosovitskiy2020vit,pmlr-v139-kim21k,beit,chen2021crossvit} usually apply \textit{[CLS]}  as an extra learnable embedding for specific tasks. However, similar to spatial patch embeddings in CNNs, the two-image-input-based model exploits the patch embedding output $\textbf{z}_L^1, \textbf{z}_L^2$ which contain information from both images, then put them into linear layers for extracting cross-image features.  We provide an ablation study to compare the performance of these cross-image features and \textit{[CLS]} in  \cref{sec:ablation}. Similar to previous deep metric learning methods in face recognition \cite{schroff2015facenet,wang2018cosface,Liu_2017_CVPR}, here we use the ArcFace as our loss function \cite{deng2018arcface} to separate and learn cross-image margins to their corresponding labels.

\subsection{Dataset}
The model is trained on the CASIA Webface \cite{yi2014learning} dataset, containing 494,414 face images of 10,575 real-world identities, widely used for FI tasks such as \cite{schroff2015facenet}. We sample 2M pairs (1M positives and 1M negatives) consisting of all identities from the processed and clean CASIA Webface dataset. 

\subsection{Evaluation against various network structures}
\label{sec:nets}

\begin{table*}[t]
\resizebox{17cm}{!}{%
\begin{tabular}{l|l|c|l|l|l|l|l}
\hline
Name  & \multicolumn{1}{c|}{Architecture} & \begin{tabular}[c]{@{}c@{}}Patch \\ Embedding\end{tabular} & \multicolumn{1}{c|}{Input} & \begin{tabular}[c]{@{}l@{}}Transformer\\ output\end{tabular} & \begin{tabular}[c]{@{}l@{}}\textbf{Inter}-image, \\ Image-wise \\ comparison\end{tabular} & \begin{tabular}[c]{@{}l@{}}\textbf{Intra}-image, \\ patch-wise \\ comparison\end{tabular} & \begin{tabular}[c]{@{}l@{}}\textbf{Inter}-image, \\ patch-wise \\ comparison\end{tabular} \\ \hline
C & CNN \cite{deng2018arcface}                       & CNN \cite{arcfacePyTorch}                                                    & 1-image                    & 1 feature                                              & \cmark                                                                                 &    Local (CNN-based)                                                                                       &  \xmark                                                                                \\ 
V & ViT \cite{dosovitskiy2020vit}                       & \emph{learned}                                                          & 1-image                    & 1 feature                                              & \cmark                                                                                 &    \cmark                                                                                       &  \xmark                                                                                \\ 
H1 & Hybrid-ViT                 & CNN                                                    & 1-image                    & 1 feature                                              &     \cmark                                                                             &    \cmark                                                                                       &  \xmark                                                                                \\ 
H2 & Hybrid-ViT                 & CNN                                                    & 2-image                    & CLS                                                    & \xmark                                                                                 &    \cmark                                                                                       &  \cmark                                                                                \\ 
H2L & {Hybrid-ViT} (\textbf{ours})                 & {CNN}                                                    & {2-image}                    & {2-Linear}                                                  &     \cmark                                                                             &    \cmark                                                                                       &  \cmark                                                                                \\ 
D & DeepFace-EMD \cite{hai2022deepface}              & CNN                                                    & 2-image                    & 2 features                                             &     \cmark  ($\alpha=0.3$)                                                                            &            Local (CNN-based)                                                                               &      \cmark  ($\alpha=0.7$)                                                                               \\ \hline
\end{tabular}
}
\caption{Properties of the six networks evaluated in this work. 
We categorize into 2 types of models: 1-image and 2-image. 
1-image models include CNN (C) and ViT (V) while the 2-image group contains DeepFace-EMD (D). 
Hybrid-ViT can be 1-image (H1) or 2-image (H2 and H2L). 
The difference between H2 and H2L is the Transformer output of \textit{[CLS]} vs. 2-Linear, respectively.}
\label{tab:networks}
\end{table*}


Here, we study six models with various architectures for face recognition, including a SNN with ArcFace \cite{deng2018arcface}, DeepFace-EMD \cite{hai2022deepface}, and Transformer ViTs, whose properties are summarized in \cref{tab:networks}.

The Siamese CNN model (denoted as C in the table) is used as a baseline in our study. The ViT-based model (denoted as V) operates at the patch level instead of the image level. The 1-image hybrid-ViT \cite{dosovitskiy2020vit} (Model H1) is the same as the original ViT except that the patch embeddings are from a pre-trained CNN, which serves as the baseline for ViT-based models. The 2-image Hybrid-ViT (Model H2) uses [CLS] for binary cross-entropy loss for one single softmax classifier layer, which we will compare to the 1-image model. The 2-image Hybrid-ViT (Model H2L) uses 2-output features for computing a cosine similarity. The 1-image model has separate ViTs for each input while the 2-image one has put two features into a single Transformer to implement cross-attention. DeepFace-EMD \cite{hai2022deepface} (D) uses entire CNN features but in two stages: First, compare images using image embeddings and then re-rank using patch embeddings. Models H2, H2L, \& D perform cross-image, patch-wise comparison---via ViT attention (H2 \& H2L) or optimal transport (D) between 2 image inputs.

For Model H2L, the spatial features embeddings (\eg $8\times8$ in ResNet-18 \cite{he2016identity}) are re-used to compute a feature vector through the linear layers which are deployed to ArcFace \cite{deng2018arcface} loss function. Utilizing this loss function for cross-image features can help transfer knowledge quickly as well as further improvements. For more details about parameter selection, see \cref{tab:networks} and \cref{sec:nets}.



\section{Ablation Studies}
\label{sec:ablation}

For model understanding and parameter selection, we conduct two major ablation studies for networks with different settings: (1) Cross-attention 2-image vs. no-cross-attention 1-image, for both in-distribution data and OOD (\cref{sec:no_cross_vs_cross_attention}), and (2) With cross-attention, 2-output linear vs. 1-output \textit{[CLS]} (\cref{sec:cls_vs_arcface}). 
In addition, we provide a study for how to select the depth and the head of Transformers (\cref{sec:effi_depth_head}).

\subsec{Datasets.} We run face verification experiments on two datasets: the in-distribution LFW \cite{yi2014learning} and the masked-face-occlusion MLFW \cite{wang2021mlfw}. The face verification task has 6,000 pairs  (3000 positives and 3000 negatives, a total of 12,000 images). For the hybrid models (C, and D), we used the pre-trained ResNet18 ArcFace model \cite{deng2018arcface}. Images are aligned and cropped to $128\times 128$ by the MTCNN algorithm \cite{schroff2015facenet}. Inputs are normalized to $[0,1]$ by subtracting $127.5$ and dividing by $127.5$. For Model V, images are cropped to $112 \times 112$ with original RGB values in $[0,255]$. All models are trained on a clean and processed CASIA Webface database \cite{yi2014learning}. 

\subsec{Model training.} We train models with a batch size of 320 images and a learning rate of $1e^{-6}$ for the first warm-up epoch and $1e^{-5}$ in the remaining 49 epochs. 
For Transformer settings, the models are trained with depth $={1,2,4, 8}$ and head $={1,2,4,6,8}$.   
For CNN backbones in hybrid-ViTs, we do not update the parameters. For ArcFace loss \cite{deng2018arcface}, hyper-parameters are mentioned in \cref{sec:training_hyperparameters}. All experiments are run on eight 40GB A100 SXM GPUs.

\subsection{The cross-attention 2-image ViT outperforms the 1-image} 

\label{sec:no_cross_vs_cross_attention}

\begin{figure*}
    \centering
    \includegraphics[width=1.0\textwidth]{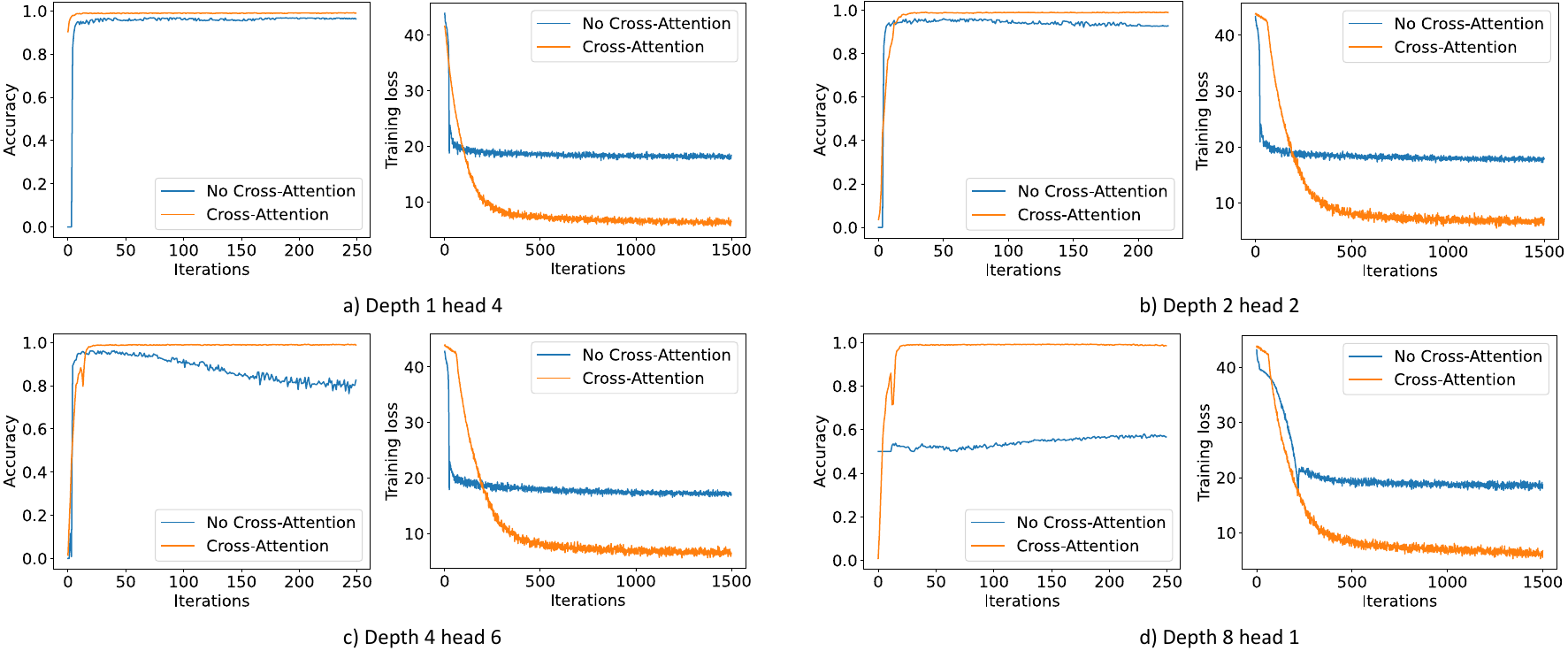}
    \caption{Comparison in accuracy and convergence between training \textcolor{RoyalBlue}{\textbf{H1}} (No-cross-attention) vs. \textcolor{orange}{\textbf{H2L}} (Cross-attention) architectures on LFW \cite{yi2014learning}. 
    For different network settings, 2-input-image achieves better accuracy and more stable training when leveraging patch-wise cross-image attention.
    } 
    \label{fig:no_cross_vs_cross}
\end{figure*}


\begin{table}
\centering
\resizebox{7cm}{!}{%
\begin{tabular}{l|l|l|l|l}
\hline
model                                                                                 & depth & head & LFW  & MLFW           \\ \hline
C~~~~~~CNN                                                                               & -     & -    & 98.02 & 70.75          \\ \hline
V~~~~~~ViT                                                                              & 20    & 8    & 97.77 & 57.62          \\ \hline
\multirow{4}{*}{\begin{tabular}[c]{@{}l@{}}H1~~~Hybrid-ViT \\ (1-image)\end{tabular}} & 1     & 4    & 96.38  & 56.00          \\ \cline{2-5} 
                                                                                      & 2     & 2    & 96.13 & 57.85          \\ \cline{2-5} 
                                                                                      & 4     & 6    & 96.20 & 57.75          \\ \cline{2-5} 
                                                                                      & 8     & 1    & 58.00 & 57.92          \\ \hline
\multirow{4}{*}{\begin{tabular}[c]{@{}l@{}}H2L Hybrid-ViT \\ (2-image)\end{tabular}} & 1     & 4    & \textbf{99.28} & \textbf{73.00} \\ \cline{2-5} 
                                                                                      & 2     & 2    & \textbf{99.27} & \textbf{71.60} \\ \cline{2-5} 
                                                                                      & 4     & 6    & \textbf{99.30} & \textbf{71.92} \\ \cline{2-5} 
                                                                                      & 8     & 1    & \textbf{99.22} & \textbf{71.90} \\ \hline
\end{tabular}
}
\caption{Comparison of 1-image (no-cross-attention) and 2-image (cross-attention). 2-image hybrid model H2L outperforms 1-image models (C, V, and H1) on in-distribution (LFW) and occlusion OOD (MLFW) domains. 
In addition, the accuracy of the low depth is similar to higher depth so that we can use the low depths. Therefore, we can rule out models: C,  V, and H1, and choose the lower depth of H2L.
}
\label{tab:no_cross_vs_cross}
\end{table}

\begin{table}
\centering
\resizebox{7cm}{!}{%
\begin{tabular}{l|l|l|l|l}
\hline
2-image Hybrid-ViT                       & depth & head & LFW  & MLFW           \\ \hline
\multirow{4}{*}{\begin{tabular}[c]{@{}l@{}}H2~~~ CLS \\ (1-output)\end{tabular}} & 1     & 1    & 90.45  & 48.40          \\ \cline{2-5} 
                            & 1     & 2    & 96.38  & 53.55          \\ \cline{2-5} 
                            & 1     & 4    & 97.47  & 56.88          \\ \cline{2-5} 
                            & 2     & 1    & 92.47  & 52.52          \\ \hline
\multirow{4}{*} {\begin{tabular}[c]{@{}l@{}}H2L~~~ 2-Linear \\ (2-output)\end{tabular}} & 1     & 1    & \textbf{99.22} &  \textbf{70.15} \\ \cline{2-5} 
                            & 1     & 2    & \textbf{99.25} & \textbf{72.77} \\ \cline{2-5} 
                            & 1     & 4    & \textbf{99.28} & \textbf{73.00} \\ \cline{2-5} 
                            & 2     & 1    & \textbf{99.28} & \textbf{70.77} \\ \hline
\end{tabular}
}
\caption{
Model H2L with 2-output features outperforms H2 (CLS output) on both LFW and MLFW.
}
\label{tab:cls_arcface}
\end{table}
To investigate our hypothesis that using cross-attention can improve the performance in face recognition, we compare our proposed 2-image (cross-attention) model with the 1-image (no-cross-attention) one. 

\subsec{Experiment.}  For Model V, we use a depth of 20 and a head of 8. For Model V \& H1, we use \textit{[CLS]} outputs to extract 512-dimension features. For Model H2L, we use the remaining 2-output with 512-dimension embeddings. All features are learned with the ArcFace loss function \cite{deng2018arcface} to classify identities.


\subsec{Results.} First, we find that the 2-image (cross-attention) model outperforms the 1-image (no-cross-attention) one significantly on the LFW and MLFW datasets, showing that cross-image information is useful for handling OOD data (\cref{tab:no_cross_vs_cross}). 
For example, in LFW, the accuracy of H2L (depth=4, head=6) increases $\sim 3.14\%$ (model H1), $\sim 1.5\%$ (Model V), and $\sim 1.25\%$ (CNN). Furthermore, the 2-image model H2L substantially provides more useful similarity information than the 1-image model for OOD distribution on MLFW (\cref{tab:no_cross_vs_cross}; Model H2L - 73\% vs. C-70.75\%, H1-57.92\%, and V-57.62\%). 

Second, interestingly, we find that the hybrid models (H1 \& H2L) can achieve higher precision with a depth of only 1, \ie adding an efficient shallow layer to Transformers can improve performance (\eg on LFW, 99.28\% H2L vs. 98.02 \% of H1). 
We deduce the same statement when comparing it with the ViT model (V).  
In contrast, the 1-image no-cross-attention model has worse performance with the in-distribution LFW (see \cref{fig:no_cross_vs_cross}) and the OOD MLFW (\cref{tab:no_cross_vs_cross}). 
With a higher depth of 8, model H1 becomes worse in LFW (\cref{tab:no_cross_vs_cross} H2L-99.22\% vs. H1-58.00\%)  

\subsection{Cross-Attention: The 2-linear-output ViT
 outperforms the 1-output [CLS]}
\label{sec:cls_vs_arcface}

\begin{figure*}
    \centering
    \includegraphics[width=1.0\textwidth]{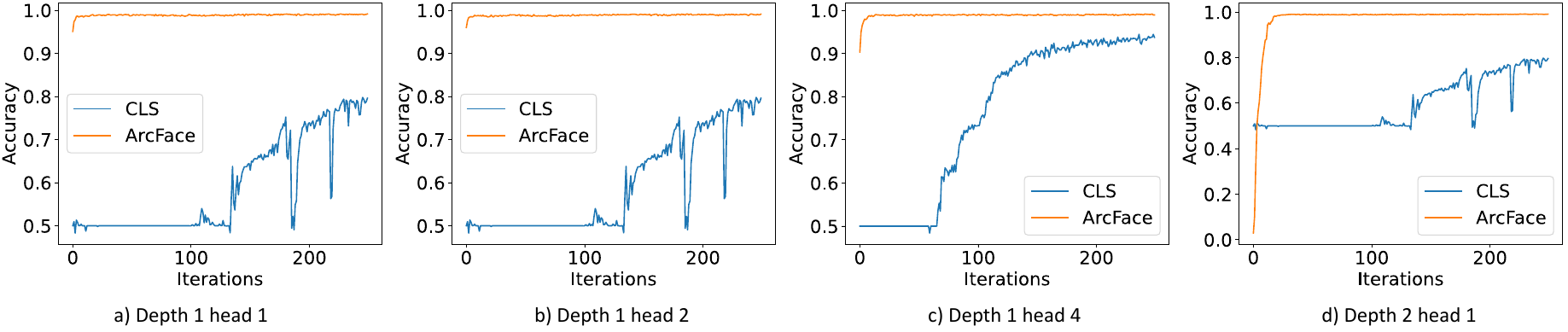}
    \caption{Training performance of CLS (model \textcolor{RoyalBlue}{\textbf{H2}}) and ArcFace hybrid-ViT (model \textcolor{orange}{\textbf{H2L}}) on LFW. Model \textcolor{orange}{\textbf{H2L}} consistently outperforms and achieves more stability in the training process.} 
    \label{fig:cls_vs_arcface}
\end{figure*}

The previous Transformer-based FI works \cite{dosovitskiy2020vit, Devlin2019BERTPO} usually use an extra learnable embedding \textit{[CLS]}, discarding the remaining embeddings that may contain helpful cross-image information. Here, we experiment with the 1-output \textit{[CLS]} (model H2) and 2-output (model H2L) to study how the embeddings can improve performance.

\subsec{Experiment.} In the 1-output \textit{[CLS]}, we deploy binary cross entropy loss to classify identities. We train Transformers with depths of 1 and 2. 

\subsec{Results.} First, we find that the 2-linear-output model H2L consistently outperforms the 1-output \textit{[CLS]} model H2 on LFW and MLFW (\cref{tab:cls_arcface}), verifying that the remaining embeddings cross-image information between two images are helpful to improve models. In LFW (in-distribution), the 2-output model improves the accuracy by \increasenoparent{8.55} points (\cref{tab:cls_arcface}; from 90.45\% of H2 to 99.22\% of H2L). In the out-of-distribution masked-face image (MLFW) datasets, the improvement is even more significant when the accuracy increases by \increasenoparent{21.75} points (\cref{tab:cls_arcface}; 48.40\% of H2 vs. 70.15\% of H2L).   

Second, the training of the 2-output Model H2L performs better and is more stable than the 1-output Model H2 in only a few iterations (\cref{fig:cls_vs_arcface}). For instance, the 1-output \textit{[CLS]} Model H2 only achieves 80\% in accuracy over LFW while the 2-output model H2L can reach 99\% in accuracy within fewer iterations (\cref{fig:cls_vs_arcface}a, b, and d).          

To sum up, we can improve model performance on OOD by using a low depth of 1, which saves computational costs and proves that H2L performs better in both in-distribution and OOD domains. In addition, with higher depths, H2 performs worse. 

\section{Main Results}
\label{sec:main_result}
In \cref{sec:face_occ,sec:face_adv},  we experiment on different OOD query types including masks, sunglasses, and adversarial faces. Here, we select the best settings from ablation studies in \cref{sec:ablation} including depth of 1 and head of 1, 2, or 6. In \cref{sec:time_complextity}, we show that our model has a faster time complexity compared with other layer types. \cref{sec:explainability} discusses our model's face explainability. To boost the performance, our proposed Model H2L can be used in a 2-stage fashion like DeepFaceEMD, \ie selecting the top 100 Stage 1's candidates ($k=100$) 
with CNN w.r.t cosine similarity scores and then re-ranking these candidates with cross-image features --- 2 outputs from Transformers. We also re-use a combination of two stages with $\alpha=0.7$, which works best for occlusion cases \cite{hai2022deepface}. The models trained with settings mentioned in \cref{sec:ablation} are reported with 2 stages (ST1 and ST2) compared with the original ArcFace and DeepFaceEMD. The results are computed by three metrics: P@1, RP, and M@R \cite{zhao2021towards,musgrave2020metric}. For the details of these metrics, see \cref{sec:eval_metrics_details}.

\subsection{Comparable accuracy}
\label{sec:face_occ}


\begin{table}
\centering
\resizebox{8.5cm}{!}{%
\begin{tabular}{c|l|c|c|c|c|c|c}
\hline
dataset                                                                       & \multicolumn{1}{l|}{name~~~~~~~~~model} & stage &depth & head & P@1   & RP    & M@R   \\ \hline
\multirow{4}{*}{\begin{tabular}[c]{@{}c@{}}CALFW  \\ (Mask)\end{tabular}}     & C~~~~~~~CNN                    & ST1 &-     & -    & 95.58 & 51.59 & 50.01 \\ 
                                                                              & H2L Hybrid-ViT               & ST1 & 1     & 2    & 95.03 & 43.70 & 42.36 \\ 
                                                                              
                                                                              & D~~~~~~~DeepFaceEMD        &ST2& -     & -    & \textbf{99.79} & \textbf{56.77} & \textbf{55.75} \\ 
                                                                              & H2L Hybrid-ViT               &ST2& 1     & 2    & 99.29 & 51.00 & 50.01 \\ \hline
                                                                              
\multirow{4}{*}{\begin{tabular}[c]{@{}c@{}}CALFW  \\ (Sunglasses)\end{tabular}} & C~~~~~~~CNN                    & ST1     & -    & -& 51.11 & 29.38 & 26.73 \\ 
                                                                              & H2L Hybrid-ViT                & ST1 & 1     & 6    & 50.23 & 28.08 & 25.15 \\ 
                                                                              & D~~~~~~~DeepFaceEMD         & ST2 & -     & -    & \textbf{54.95} & 30.66 & 27.74 \\ 
                                                                              & H2L Hybrid-ViT (ST2)               & ST2 & 1     & 6    & 54.00 & \textbf{31.00} & \textbf{27.87} \\ \hline
\multirow{4}{*}{\begin{tabular}[c]{@{}c@{}}AgeDB  \\ (Mask)\end{tabular}}     & C~~~~~~~CNN                   & ST1     & -    & - & 96.31 & 39.22 & 30.41 \\ 
                                                                              & H2L Hybrid-ViT                & ST1 & 1     & 1    & {98.73} & 20.68 & 14.86 \\ 
                                                                              & D~~~~~~~DeepFaceEMD       & ST2 & -     & -    & \textbf{99.84} & \textbf{39.22} & \textbf{33.18} \\ 
                                                                              & H2L Hybrid-ViT                 & ST2 & 1     & 1    & 99.28 & 33.93 & 26.69 \\ \hline
\multirow{4}{*}{\begin{tabular}[c]{@{}c@{}}AgeDB  \\ (Sunglasses)\end{tabular}} & C~~~~~~~CNN                    & ST1     & -    & - & 84.64 & 51.16 & 45.00 \\ 
                                                                              & H2L Hybrid-ViT                 & ST1 & 1     & 2    & 86.01 & 49.34 & 43.03 \\ 
                                                                              & D~~~~~~~DeepFaceEMD      & ST2& -     & -    & \textbf{87.06} & 50.04 & 44.27 \\ 
                                                                              & H2L Hybrid-ViT               & ST2 & 1     & 2    & 86.75 & \textbf{51.16} & \textbf{44.88} \\ \hline
\multirow{4}{*}{\begin{tabular}[c]{@{}c@{}}TALFW  \\ vs.\\ LFW\end{tabular}}     & C~~~~~~~CNN                   & ST1 & -     & -    & 93.49 & 81.04 & 80.35 \\ 
                                                                              & H2L Hybrid-ViT                & ST1 & 1     & 2    & 94.59 & 77.66 & 77.00 \\ 
                                                                              & D~~~~~~~DeepFaceEMD        & ST2 & -     & -    & \textbf{96.64} & \textbf{82.72} & \textbf{82.10} \\ 
                                                                              & H2L Hybrid-ViT             & ST2 & 1     & 2    & 94.03 & 81.63 & 81.09 \\ \hline
\end{tabular}
}
\caption{Face occlusions and adversarial images. Model H2L achieves comparable accuracy on the OOD of CALFW and AgeDB compared to CNN and DeepFace-EMD \cite{hai2022deepface}.
}
\label{tab:face_occ}
\end{table}

\subsec{Experiment.} We demonstrate our models for FI on two datasets: CALFW \cite{Tianyue2017calfw} and AgeDB \cite{moschoglou2017agedb}. The 12,173 CALFW images and 16,488 AgeDB images have age-varying of 4,025 and 568 identities, respectively. We re-use OOD queries of these datasets from DeepFaceEMD \cite{hai2022deepface} consisting of masks and sunglasses.

\subsec{Results.}First, in ST1, 2-image (model H2L) achieves comparable accuracy with the original ArcFace \cite{deng2018arcface}. 
In the AgeDB dataset, ST1's P@1 of model H2L improves around \increasenoparent{2} points over model C on Mask (98.73\% vs. 96.31\%; \cref{tab:face_occ}) and Sunglasses (86.01\% vs. 84.64\%; \cref{tab:face_occ}), increasing the accuracy on occlusion in the cross-age domain. 

Second, ST2 of Model H2L significantly outperforms ST1 (\eg CALFW (mask) \textbf{99.29}\% vs. \textbf{95.58}\% P@1 in \cref{tab:face_occ})
and achieves better results compared with DeepFaceEMD in sunglass images (ST2 on RP and M@R metrics in \cref{tab:face_occ}), verifying the boost performance in the 2-stage process.

\subsection{Comparable robustness}
\label{sec:face_adv}

\subsec{Experiment.} To illustrate the effectiveness of adversarial attacks, we run the experiment on the TALFW dataset \cite{zhong2020towards}. TALFW contains 13,233 images perturbed adversarially to fool face models. 

\subsec{Results.} First, in ST2, model H2L achieves better results than model H1 on all 3 metrics, P@1 (H2L-94.03\% vs. H1-93.49\%), RP (H2L-81.63\% vs. H1-81.04\%), and M@R (H2L-81.09\% vs. H1-80.35\%). 
See the last row of \cref{tab:face_occ}), verifying that our proposed model H2L also improves the precision in adversarial images with a re-ranking algorithm.
Second, DeepFace-EMD (model D) achieves the best results in all metrics both ST1 and ST2 (see the last row of \cref{tab:face_occ}). 
These results show that these models (models H2L \& D) 
are robust to adversarial images, which is a grand challenge in computer vision \cite{kurakin2016adversarial,nguyen2015deep}.  

\subsection{Faster inference time}
\label{sec:time_complextity}

\begin{table}[h!]
\centering
\resizebox{8cm}{!}{%
\begin{tabular}{l|l|l|l}
\hline
Layer type                                                                                        & \begin{tabular}[c]{@{}l@{}}Complexity \\ per layer\end{tabular} & \begin{tabular}[c]{@{}l@{}}Actual \\ runtime \\ (s)\end{tabular} & \begin{tabular}[c]{@{}l@{}}Maximum \\ path \\ Length\end{tabular} \\ \hline
C. Convolutional                                                   & $O(k \cdot n \cdot d^2)$                                                                &  -                                                                &    $O(\log_kn)$                                                            \\ 
V. ViT, Self-Attention                                                       &   $O(n^2 \cdot d)$                                                              &  -                                                                &     $O(1)$                                                           \\ 

V. Self-Attention (restricted)                                                                       & $O(r \cdot n \cdot d^2)$                                                                & -                                                                 &    $O(n/r)$                                                            \\ 
H2L Hybrid-ViT &     $O(k\cdot n \cdot d^2 + n^2 \cdot d)$                                                            &   \textbf{24.33}                                                               &  $O(\log_k n)$                                                               \\ 
D. DeepFace-EMD \cite{hai2022deepface}                                                                                           & $O(k \cdot n \cdot d^2 + n^3 \cdot \log n)$ \cite{Shirdhonkar2008Cvpr}                                                                &   53.35                                                               &    $O(1)$                                                            \\ 

\hline
\end{tabular}
}
\caption{Time complexity of different type layers. n is the number of patches, d is the dimension of embeddings, k is the kernel size of convolutions, and r is the size of the neighborhood in restricted self-attention.}

\label{tab:time_com}
\end{table}

The run-time complexities of Model C, V, H2L, and D are shown in \cref{tab:time_com} and detailed in \cref{sec:time_com}. Our Model H2L has a lower complexity, $O(n^2)$, than that of DeepFace-EMD, $O(n^3)$. In practice, Model H2L performs at least 2 times faster than Model D when used as the re-ranking process (ST2) in face identification (see \cref{tab:time_run}, \cref{fig:model_inference_time_on_ood}). Moreover, in ST2, DeepFace-EMD is slow to solve EMD for higher dimension patch-wise similarity \cite{hai2022deepface} while hybrid-ViT simply computes the cosine similarity of cross-image features and low-depth Transformers, \ie enhancing the scalability. For example, in AgeDB (sunglasses), the computation is sped up to $\sim\textbf{3}\times$ for 16,409 sunglass-query images in settings of $8\times8$ patches (see \cref{tab:time_run} for details).
Therefore, model H2L is a good choice for more scalable architectures. 

\begin{figure*}
\begin{flushleft}
    \hspace{2cm}
    (C) CNN     \hspace{1.5cm} 
    (V) ViT \hspace{1.2cm}
    (H1) ViT-attn \hspace{0.5cm}
    (H2L) Hybrid-ViT \hspace{0.8cm}
    (D) DeepFace-EMD \hfill
    \end{flushleft}
    \centering
    \vspace{-0.3cm}
    \includegraphics[width=1.0\textwidth]{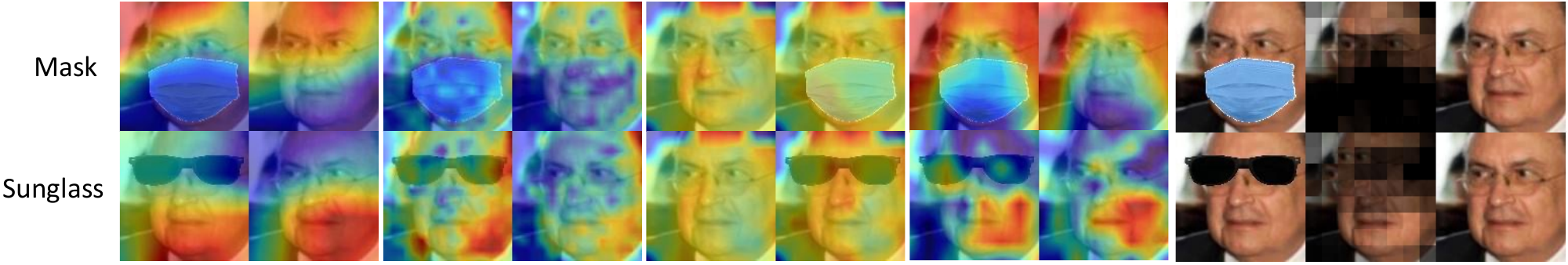}
    \caption{
    Comparison of face models' explainability on \textbf{LFW} OOD domains. ViT-attn is visualized through the method of Chefer et al. \cite{chefer2021generic}. 
    Our proposed H2L can highlight the important area in images (\eg eyes, mouth, etc.) and remove occluded areas (\eg mask and sunglasses). 
    In contrast, Model V contains noisy heatmaps and H1 does not provide any interpretable clues of how two faces match.  
    } 
    \label{fig:explain_vis}
\end{figure*}

\begin{figure}[ht]
    \centering
    \includegraphics[width=0.5\textwidth]{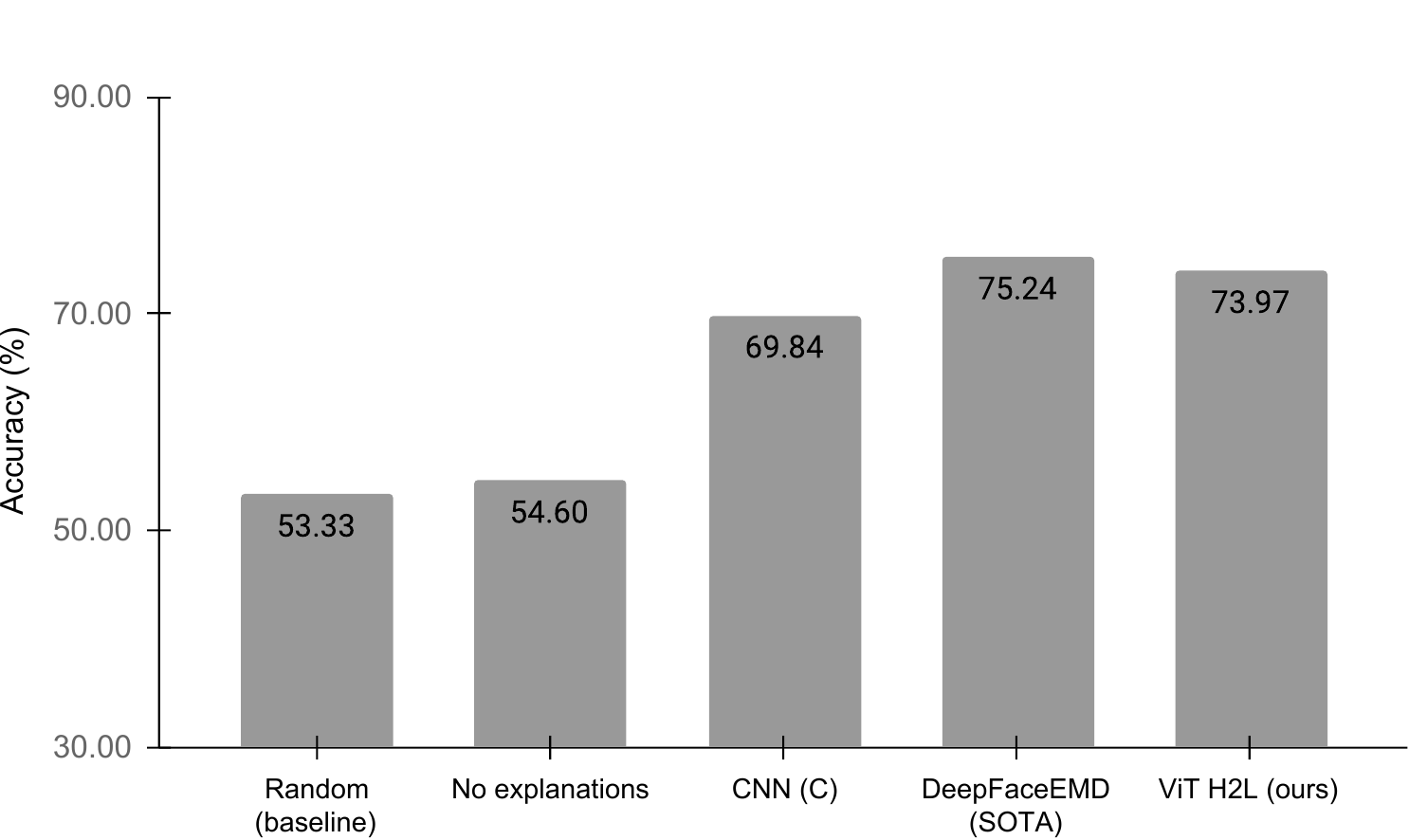}
    \caption{Human explainability across various networks. The mean and standard deviation of the accuracy of 21 users when presented with 4 explanations: Cross-correlation (CC) method on CNNs \cite{stylianou2019visualizing}; flow visualization in DeepFace-EMD \cite{hai2022deepface}; CC on 2-image Hybrid-ViT; and a baseline of no explanations. The explanations of Model D and H2L result in substantially higher user accuracy than those of Model C and the No-explanation baseline, which is close to the random baseline of 53.33\%.}
    \label{fig:human_explainability_across_networks}
\end{figure}

\subsection{Better model explanation by human evaluation}
\label{sec:explainability}

As face identification systems in the real world are often customer-facing \cite{lawsuit2021facial,newjersey2021arrested,detroit2021arrested,michigan2021arrested,MiaSato.2021}, we study how CNNs (model C), 1-image ViTs (model V), 2-image Hybrid-ViT (model H2L), and DeepFace-EMD (model D) help users in understanding face verification results. For each image pair, we generate a visual explanation from a model (examples in \cref{fig:explain_vis}), and ask a user to look at both images and the explanation and decide whether the two faces are of the same person.

\subsec{Experiment.}  Similar to \cite{zhao2021towards,hai2022deepface}, we use the cross correlation method from \cite{stylianou2019visualizing} to generate similarity heatmaps for the CNNs and ViTs. This method produces a heatmap by taking the dot product between every patch embedding of image 1 and the global average pooling feature of image 2. For DeepFace-EMD, we plot their flow visualizations as in \cite{hai2022deepface}.


The explanation heatmaps are generated for models C, H2L, and D using their last convolutional layers, which have the same spatial dimension of 8$\times$8.
For model V, the spatial dimension of the heatmap is 14$\times$14.
In preliminary experiments, we find the raw cross-attention matrices at the first layer of the ViT model uninformative to users (see \cref{fig:explain_vis}; ViT-attn).
Therefore, we use cross-correlation (CC) \cite{stylianou2019visualizing} to generate explanations for ViTs (\cref{fig:explain_vis}; ViT).


We recruit 21 participants who are graduate students across multiple institutions in the U.S., Vietnam, and China. 
For each user, we provide them 5 training examples and 15 pairs of images per method (\ie 15 pairs $\times$ 4 methods = 60 pairs in total).
We randomly mask and place a pair of sunglasses on each image. ~\cref{sec:study_samples} presents specific examples and how we design for user study.

\subsec{Results.}
First, we find that users without any model explanations score an average accuracy of 54.60\%, \ie near random chance (53.33\%).
This suggests that the face verification task is challenging to users (which is consistent with the qualitative feedback obtained from users).

Second, all model explanations are useful in improving user accuracy.
Model H2L and D are most useful to users who score 73.97\% and 75.24\% respectively.
Interestingly, these explanations of Model H2L and D, which leverage cross-image interaction, are more useful than the CC explanations of CNNs, which do not allow cross-image interaction (69.84\% user accuracy; \cref{fig:human_explainability_across_networks}).
In sum, consistent with the accuracy-based analysis in \cref{sec:face_occ} \& \cref{sec:face_adv}, our user study finds models with cross-image interaction (Model H2L and F) have higher explainability to users.

\section{Discussion and Conclusion}
\label{sec:con}


First, we find that using models that leverage cross-image interaction as the re-ranker substantially improves FI accuracy under occlusion and adversarially perturbed queries. Second, we train a 2-image Hybrid-ViT model that not only achieves similar accuracy but also two times faster than state-of-the-art models. Note that the 1-image models remain the fastest due to efficient image embedding caching. Finally,  visual explanations in cross-image interaction models greatly enhance lay-user face verification accuracy. We also conduct the inaugural study comparing state-of-the-art FI approaches based on accuracy, complexity, and explainability.

\subsec{Significance.} Face identification in the wild is essentially a hard, ill-posed zero-shot image retrieval task. We hope our work can inspire more explorations in the use of ViTs for face identification and to improve the speed of this system in the real world.


\subsec{Future work.} The performance of hybrid-ViTs is still slightly lower than that of DeepFace-EMD. 
It would be possible to tune ViT hyperparameters \cite{Beyer2022vit_baseline} for higher accuracy and incorporate sparsity into the attention mechanism of ViT for improved inference speed.

\subsec{Acknowledgement}
We thank Thang Pham, Giang Nguyen, Qi Li, Peijie Chen, Mohammad Reza Taesiri, and anonymous reviewers of WACV 2024 for valuable suggestions. 
Thanks to the 21 participants who dedicated their time to complete our user experiments. 
We are also grateful to AL EPSCoR GRSP 17 for funding HP. 
AN received support from NSF Grant No. 1850117 and a donation from Adobe Research and the NaphCare Foundation.



\clearpage
\newpage

{\small
\bibliographystyle{ieee_fullname}
\bibliography{egbib}
}

\newpage
\clearpage

\renewcommand{\thesection}{S\arabic{section}}
\renewcommand{\thesubsection}{\thesection.\arabic{subsection}}

\newcommand{\beginsupplementary}{%
            \setcounter{table}{0}
    \renewcommand{\thetable}{S\arabic{table}}%
            \setcounter{figure}{0}
    \renewcommand{\thefigure}{S\arabic{figure}}%
    \setcounter{section}{0}
}
\newcommand{\suptitle}{Appendix for:\\\papertitle}

\newcommand{\toptitlebar}{
    \hrule height 4pt
    \vskip 0.25in
    \vskip -\parskip%
}
\newcommand{\bottomtitlebar}{
    \vskip 0.29in
    \vskip -\parskip%
    \hrule height 1pt
    \vskip 0.09in%
}

\beginsupplementary%

\newcommand{\maketitlesupp}{
    \onecolumn
    \begin{@twocolumnfalse}
        \null%
        \vskip .375in
        \begin{center}
            {\Large \bf \suptitle\par}
            \vspace*{24pt}
            {
                \large
                \lineskip=.5em
                \par
            }
            \vskip .5em
            \vspace*{12pt}
        \end{center}
    \end{@twocolumnfalse}
}

\maketitlesupp

\section{Pre-trained models}
\label{sec:supp_pretrained_models}

\paragraph{Sources} We downloaded the three pre-trained PyTorch models of ArcFace from:

\begin{itemize}
    \item ArcFace \cite{deng2018arcface}: \url{https://github.com/ronghuaiyang/arcface-pytorch}
    
\end{itemize}

ArcFace models were trained on dataset CASIA Webface \cite{yi2014learning}.

\paragraph{Architectures} The network architectures are provided here:

\begin{itemize}
    \item ArcFace: \url{https://github.com/ronghuaiyang/arcface-pytorch/blob/master/models/resnet.py}
\end{itemize}

\paragraph{Image-level embeddings for Ranking} We use the following layer to extract the image embeddings for stage 1, \ie, ranking images based on the cosine similarity between each pair of (query image, gallery image).

\begin{itemize}
    \item Arcface: layer~\layer{bn5} (see \href{https://github.com/ronghuaiyang/arcface-pytorch/blob/master/models/resnet.py#L177}{code}), which is the 512-output, last BatchNorm linear layer of ArcFace (a modified ResNet-18 \cite{he2016identity}).
\end{itemize}

\paragraph{Patch-level embeddings for Re-ranking}

We use the following layer to extract the spatial feature maps (\ie embeddings $\{q_i\}$) for the patches:

\begin{itemize}
    \item ArcFace: layer \layer{dropout} (see \href{https://github.com/ronghuaiyang/arcface-pytorch/blob/master/models/resnet.py#L175}{code}). Spatial dimension: $8 \times 8$.
\end{itemize}

\section{Training hyperparameters}
\label{sec:training_hyperparameters}
Here we describe the hyperparameters used for ArcFace as follows.
\begin{itemize}
    \item Margin: $m=0.5$
    \item Feature scale: $s=30.0$
\end{itemize}

\section{Evaluation metrics}
\label{sec:eval_metrics_details}
P@1 is well-known as Recall@1 in metric learning. 
P@1 is computed as follow.
$$
\mathcal{N}_{q}^{k}=\underset{\mathcal{N} \subset \mathcal{X}_{\text {test }},|\mathcal{N}|=k}{\arg \min } \sum_{x^{f} \in \mathcal{N}} d_{e}\left(\phi\left(x^{q}\right), \phi\left(x^{f}\right)\right)
$$
where $x^{q}$ and $\phi(\cdot)$ are inputs and feature encoder respectively, $d_{e}(\cdot, \cdot)$ is the euclidean distance, and $k$ is $k$-nearest neighbors. Precision@k can be calculated as:
$$
\mathrm{P} @ k=\frac{1}{\left|\mathcal{X}_{\text {test }}\right|} \sum_{x_{q} \in \mathcal{X}_{\text {test }}} \frac{1}{k} \sum_{x^{i} \in \mathcal{N}_{q}^{k}} \begin{cases}1, & y^{i}=y^{q} \\ 0, & \text { otherwise }\end{cases}
$$
where $y^{i}$ is the class label of sample $x^{i}$. 


To gain more information and a comprehensive ranking evaluation, we computed mean average precision of R (M@R \cite{musgrave2020metric}), where R is number of images in a class.
$$
M@R=\frac{1}{R} \sum_{i=1}^{R} P(i)
$$
where
$$
P(i)= \begin{cases}\mathrm{P} @ i, & \text { if the } i \text {-th retrieval is correct; } \\ 0, & \text { otherwise. }\end{cases}
$$


\section{Time complexity}
\label{sec:time_com}
Here, we evaluate the time complexity of different layers. For vanilla ViT and CNN, the time complexity is mentioned in the Transformer network \cite{vaswani2017attention} (see  \cref{tab:time_com}). Hybrid-ViTs consist of convolutional neural networks and low-depth self-attention at the top.  Hence, time complexity can be added by convolutional layers and self-attention layers (the last row in \cref{tab:time_com}).

\section{The low depth's efficiency}
\label{sec:effi_depth_head}
\begin{figure*}
\begin{flushleft}
    \hspace{1.5cm}
    (a) LFW     \hspace{2.2cm} 
    (b) TALFW \hspace{2.3cm}
    (c) MLFW \hfill
    \end{flushleft}
    \includegraphics[width=1.0\textwidth]{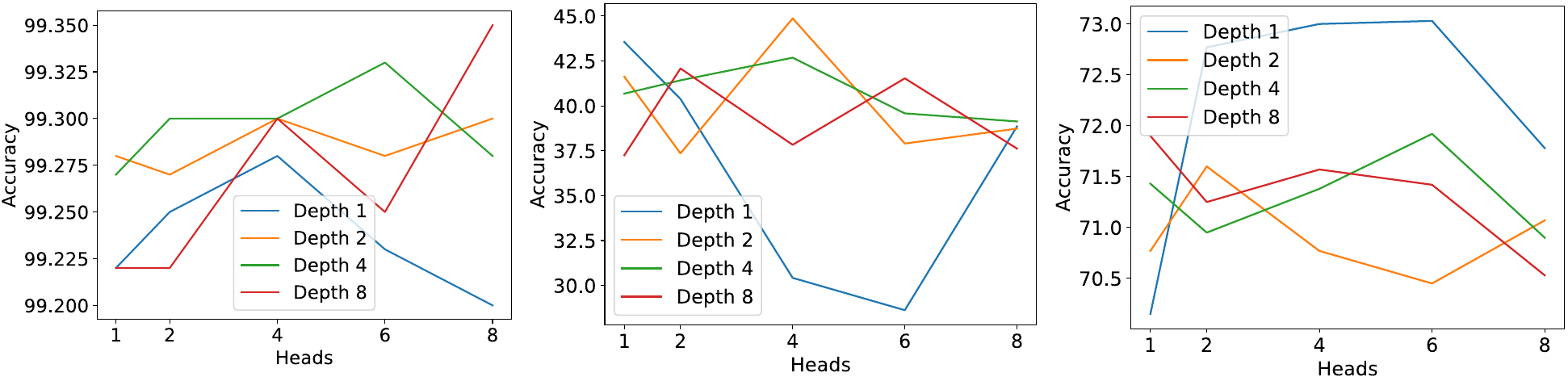}
    \caption{The efficiency of settings of depths and heads for the network (\textbf{H2L}) within different domains. For LFW, the depth of 1 achieved comparable accuracy with a depth of 8 (\eg very small difference of ~0.075 \%). In TALFW, with depths of 1 and 2 and heads of 1 and 4 respectively, the accuracy outperforms the accuracy of depths of 4 and 8. For face masks in MLFW, the depth of 1 consistently outperforms the other settings. Therefore, using a low depth of \textbf{1} or \textbf{2} for contextual information design can gain good performance.
    }
    \label{fig:effi_depth_head}
\end{figure*}

For the 2-image/output hybrid-ViT ({H2L}), adding the Transformer layer at the top of CNN can improve the performance. However, increasing the number of depths and heads can lead to redundant computation in the models while showing no meager improvements in terms of accuracy. Here, we evaluate the effects of different values of depths and heads to select the potential settings in face problems.

\subsec{Experiment} As mentioned above, we use  $\text{depths}=1,2,4,8$ and  $ \text{heads} = 1,2,4,6,8$. 
We report the accuracy of H2L for face verification on LFW, TALFW, and MLFW datasets.  

\subsec{Results} First, we observe that on LFW, the low depth of 1 achieves lower performance. However, it still \textbf{outperforms the CNN model} (99.28\%  \cref{fig:effi_depth_head} (head=4) vs 98.02\% in \cref{fig:no_cross_vs_cross}). Moreover, a lower value of depth and head \textbf{achieves comparable results} compared to higher values (\eg 99.22\% with depth of 1, head of 1, vs. 99.34\% with depth of 8, head of 8). Second, for the TALFW dataset, hybrid-ViTs (H2L) also \textbf{achieves comparable accuracy}  (with $\text{d}$=1, $\text{h}$=1), \ie performing well with low depth and head values for face adversarial. Third, for the MLFW dataset, the depth of 1 \textbf{outperforms} the other higher-depth value models. 

\section{User study samples}
\label{sec:study_samples}
\begin{figure*}[!ht]
    \centering
    \includegraphics[width=0.9\textwidth]{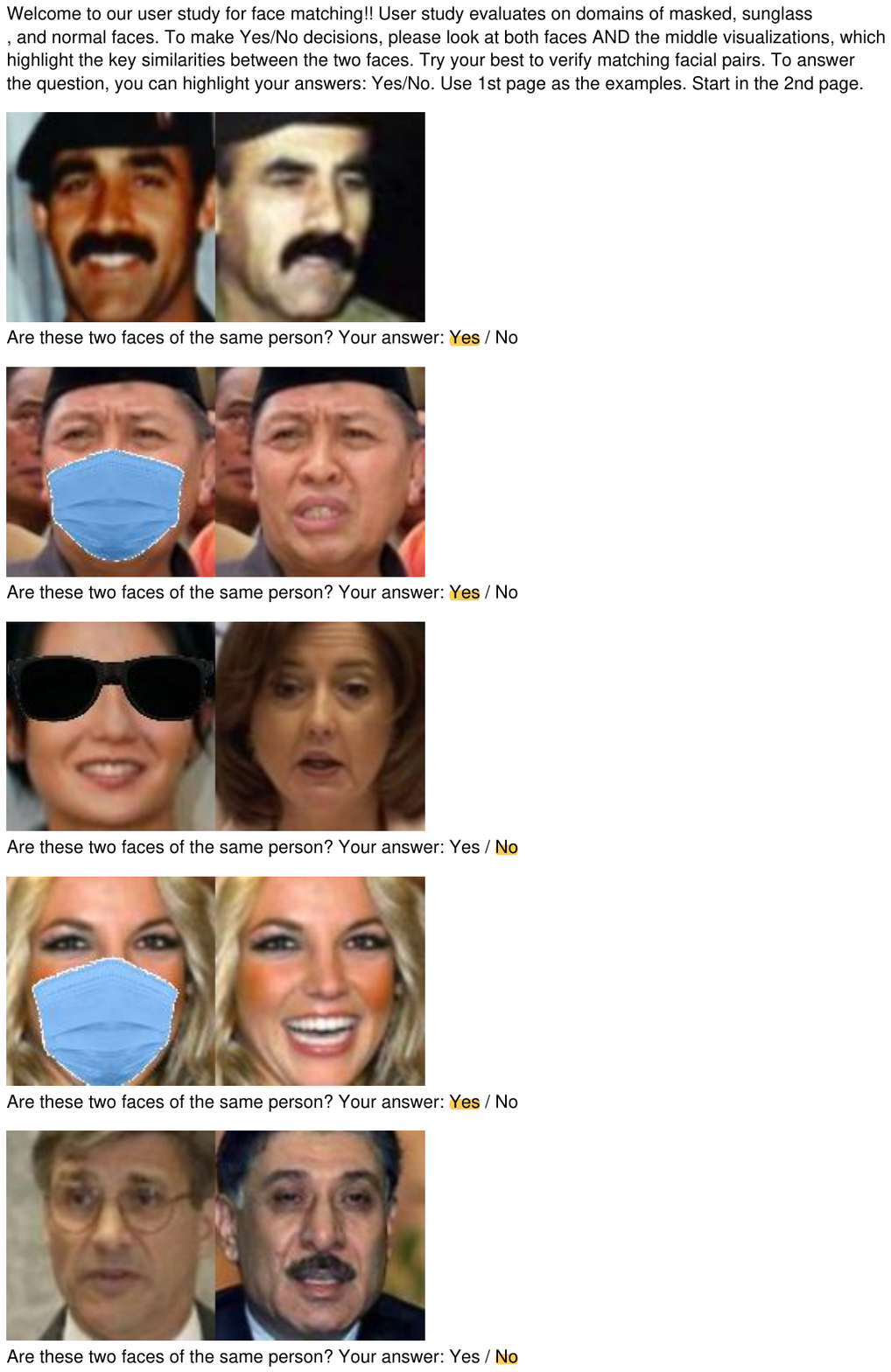}
    \caption{ User study for no-explanation method.
    }
    \label{fig:no_ex}
\end{figure*}

\begin{figure*}[h!]
    \centering
    \includegraphics[width=0.9\textwidth]{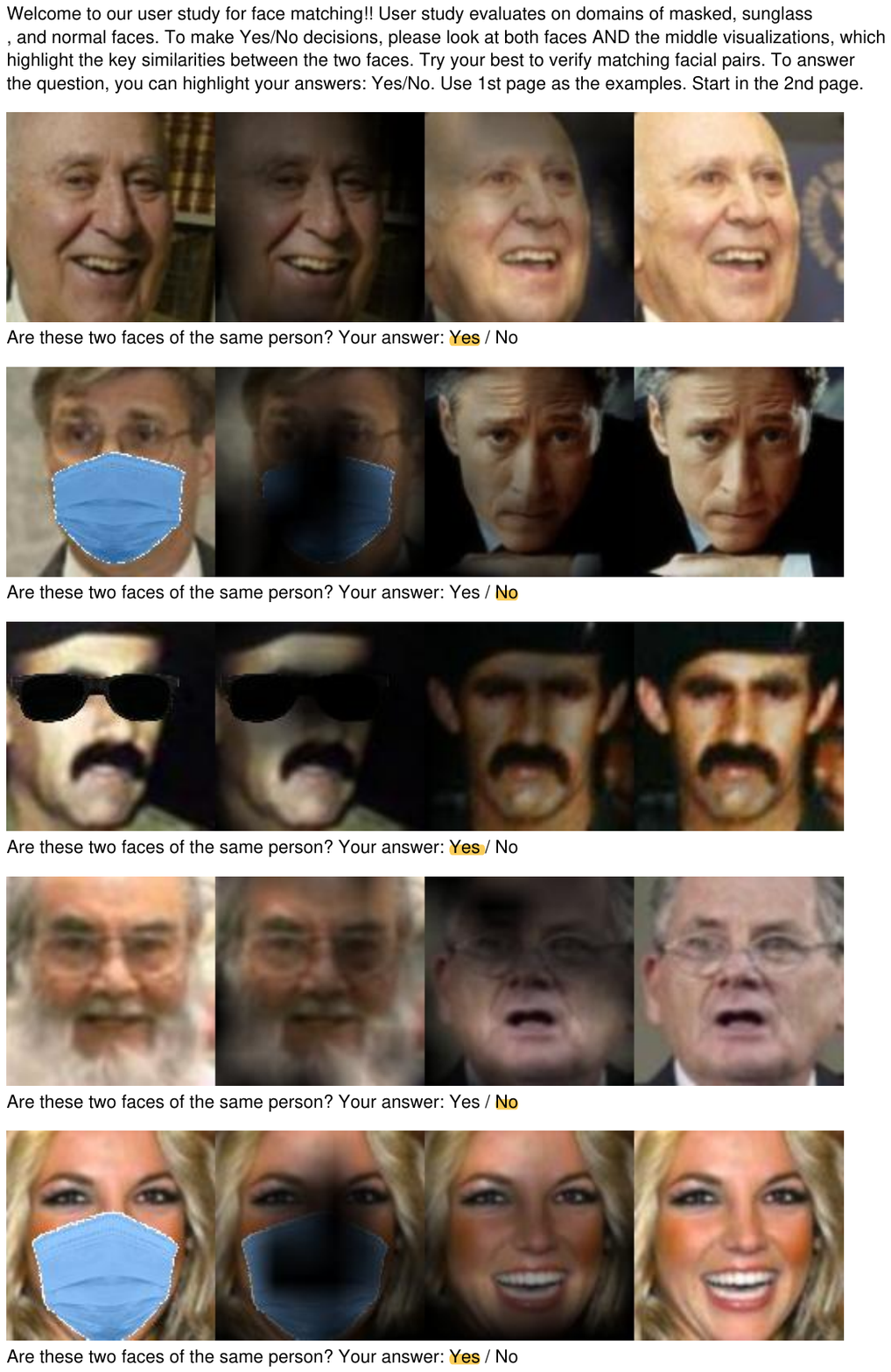}
    \caption{ User study for CNNs method.
    }
    \label{fig:cc_ex}
\end{figure*}

\begin{figure*}[h!]
    \centering
    \includegraphics[width=0.9\textwidth]{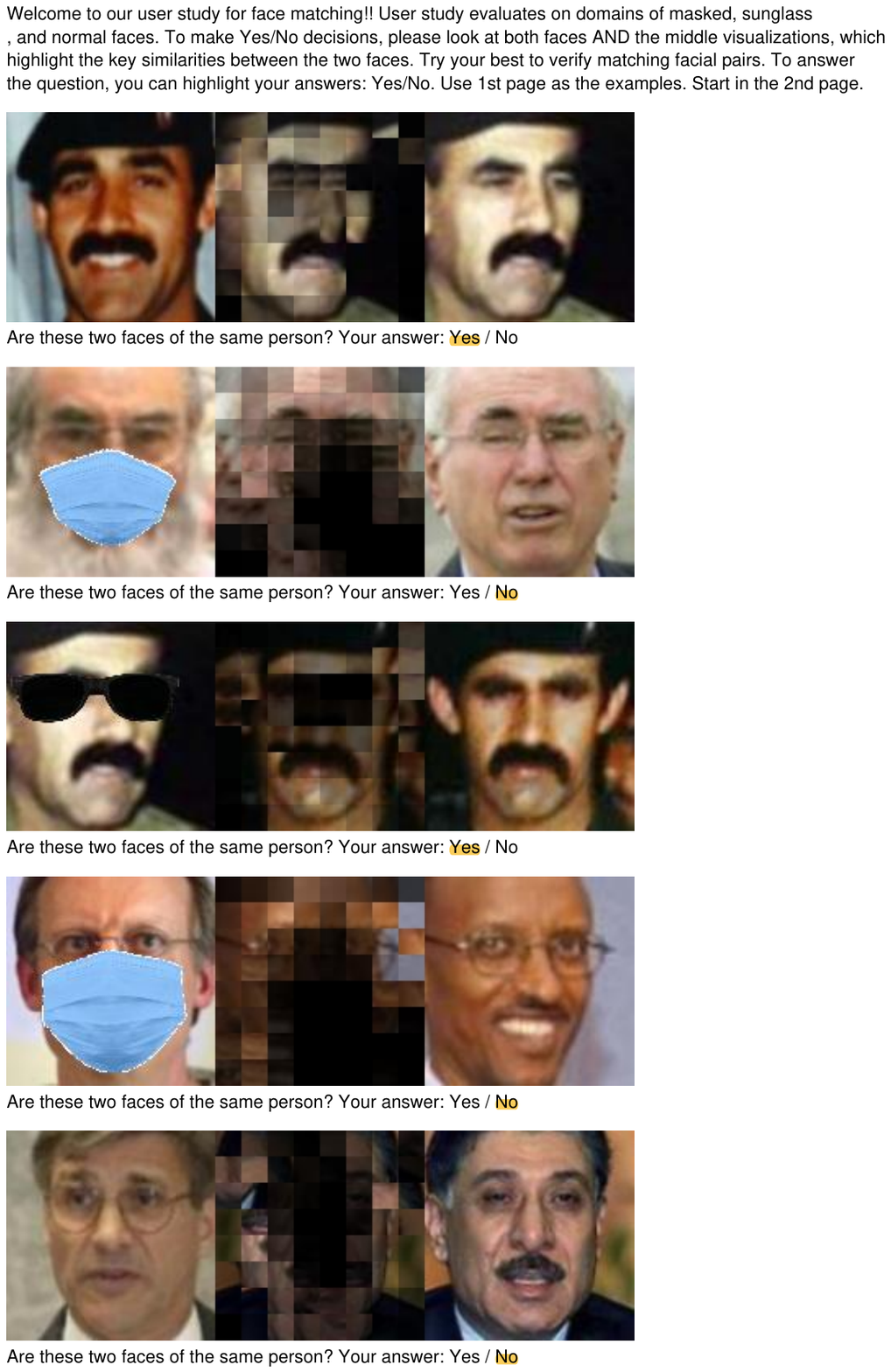}
    \caption{ User study for the EMD method.
    }
    \label{fig:emd_ex}
\end{figure*}

~\cref{fig:no_ex}, ~\cref{fig:cc_ex}, ~\cref{fig:cc_ex}, and ~\cref{fig:emd_ex} are specific examples for our design for user study. These figures are only the first pages to instruct users for each approach.

\begin{figure*}[h!]
    \centering
    \includegraphics[width=0.9\textwidth]{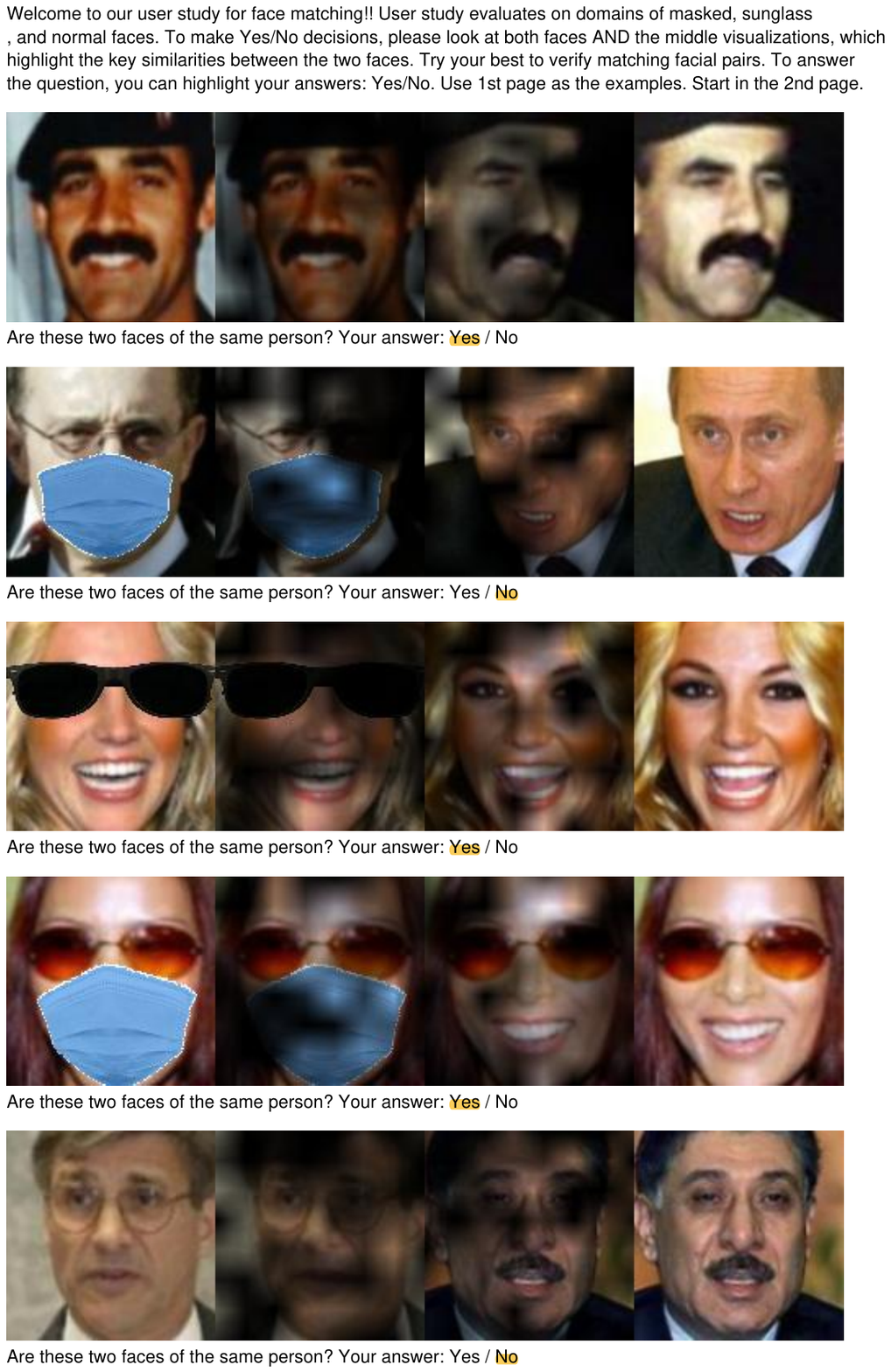}
    \caption{ User study for Hybrid-ViT method.
    }
    \label{fig:vit_ex}
\end{figure*}

Here, we experiment with 4 approaches: no explanation, Hybrid-ViT, CNNs \cite{stylianou2019visualizing}, and EMD \cite{hai2022deepface}.
\section{Architectures}
\label{sec:arch}
We provide code snippets for architectures of Transformers, H2L, and H2.
\begin{python}
class Transformer(nn.Module):
    def __init__(self, dim, depth, heads, dim_head, mlp_dim, dropout):
        super().__init__()
        self.layers = nn.ModuleList([])
        for _ in range(depth):
            self.layers.append(nn.ModuleList([
                Residual(PreNorm(dim, Attention(dim, heads=heads, dim_head=dim_head, dropout=dropout))),
                Residual(PreNorm(dim, FeedForward(dim, mlp_dim, dropout=dropout)))
            ]))
    def forward(self, x, mask=None):
        attns = []
        for attn, ff in self.layers:
            att, x = attn(x, mask=mask)
            if att is not None:
                attns.append(att)
            _, x = ff(x)
        return x, attns
        
class H2L(nn.Module):
    def __init__(self, *, num_class, image_size, patch_size, ac_patch_size,
                         pad, dim, depth, heads, mlp_dim, resnet_model, channels=3, 
                         dim_head=64, dropout=0., emb_dropout=0., out_dim=512):
        super().__init__()
        num_patches = 8 ** 2 
        patch_dim = channels * ac_patch_size ** 2
        self.resnet_model = resnet_model  # resnet to extract embeedings
        self.sep = nn.Parameter(torch.randn(1, 1, dim))    
        self.pos_embedding = nn.Parameter(torch.randn(1, 2*num_patches + 2, dim))
        self.cls_token = nn.Parameter(torch.randn(1, 1, dim))
        self.dropout = nn.Dropout(emb_dropout)
        self.transformer = Transformer(dim, depth, heads, dim_head, mlp_dim, dropout)
        self.to_latent = nn.Identity()
        self.ln = nn.LayerNorm(out_dim)  # Layer norm
        self.soft_split = nn.Unfold(kernel_size=(ac_patch_size, ac_patch_size),
                                    stride=(self.patch_size, self.patch_size),
                                    padding=(pad, pad)
                                    )
        self.patch_to_embedding = nn.Linear(patch_dim, dim)
        self.bn1 = nn.BatchNorm1d(out_dim)
        self.bn2 = nn.BatchNorm1d(out_dim)
        self.fc1 = nn.Linear(d*d*dim, out_dim)
        self.fc2 = nn.Linear(d*d*dim, out_dim)
        self.loss = ArcFace(in_features=out_dim, out_features=num_class)

    def forward(self, img, label=None, mask=None):
        out = self.resnet_model(img)
        x = out['embedding_88']
        N, C, _, _ = x.size()
        x = x.view(N, C, -1).transpose(1, 2)
        b, n, _ = x.shape
        half = int(N/2)
        
        cls_tokens = repeat(self.cls_token, '() n d -> b n d', b = int(b/2))
        
        sep = repeat(self.sep, '() n d -> b n d', b=int(b/2))
        splits = torch.split(x, half)
        x = torch.cat((splits[0], sep, splits[1]), dim=1)
        
        x = torch.cat((cls_tokens, x), dim=1)
        x = self.dropout(x)

        x, attns = self.transformer(x, mask)
        N, d, C = x.size()

        half = int(d/2)
        x1, x2 = x[:, 1:half, :], x[:, (half + 1):d, :]
        embedding1, embedding2 = x1, x2
        f1 = x1.mean(dim=1) 
        f2 = x2.mean(dim=1)

        x1 = x1.view(x1.size(0), -1)
        x2 = x2.view(x2.size(0), -1)

        x1 = self.fc1(x1)
        x1 = self.bn1(x1)
        x1 = self.ln(x1)
        
        x2 = self.fc2(x2)
        x2 = self.bn2(x2)
        x2 = self.ln(x2)

        x = torch.cat((x1, x2), dim=0)
        x = self.loss(x, label)
        return x    

class H2(nn.Module):
    def __init__(self, *, num_class, image_size, patch_size, ac_patch_size,
                          pad, dim, depth, heads, mlp_dim, resnet_model, channels=3, 
                          dim_head=64, dropout=0., emb_dropout=0., out_dim=512):
        super().__init__()
        num_patches = 8 ** 2
        patch_dim = channels * ac_patch_size ** 2
        self.patch_size = patch_size
        self.resnet_model = resnet_model # # resnet to extract embeedings
        self.sep = nn.Parameter(torch.randn(1, 1, dim))
        self.pos_embedding = nn.Parameter(torch.randn(1, 2*num_patches + 2, dim))
        self.cls_token = nn.Parameter(torch.randn(1, 1, dim))
        self.dropout = nn.Dropout(emb_dropout)
        self.transformer = Transformer(dim, depth, heads, dim_head, mlp_dim, dropout)
        self.to_latent = nn.Identity() 
        self.ln = nn.LayerNorm(out_dim)
        self.fc = nn.Linear(dim, 2)  # outputs

    def forward(self, img, label=None, mask=None):
        if self.face_model:
        out = self.resnet_model(img)
        x = out['embedding_88']
        N, C, _, _ = x.size()
        x = x.view(N, C, -1).transpose(1, 2)
        b, n, _ = x.shape
        half = int(N/2)
        cls_tokens = repeat(self.cls_token, '() n d -> b n d', b = int(b/2))
        sep = repeat(self.sep, '() n d -> b n d', b=int(b/2))
        splits = torch.split(x, half)
        x = torch.cat((splits[0], sep, splits[1]), dim=1)
        x = torch.cat((cls_tokens, x), dim=1)
        x += self.pos_embedding[:, :(2*n + 2)]
        x = self.dropout(x)
        x, attns = self.transformer(x, mask)
        N, d, C = x.size()
        x = x[:, 0, :]
        x = self.ln(x)
        x = self.fc(x)
        return x
\end{python}

\begin{table*}
\centering
\resizebox{14cm}{!}{%
\begin{tabular}{c|l|c|c|c|c|c|c|c}
\hline
Dataset                                                                       & \multicolumn{1}{c|}{Model} &\# of queries & Time (seconds) &Depth & Head & P@1   & RP    & M@R   \\ \hline
\multirow{2}{*}{\begin{tabular}[c]{@{}c@{}}CALFW  \\ (Mask)\end{tabular}}     & D~~~~~~~DeepFaceEMD  \cite{hai2022deepface}         & \multirow{2}{*}{\begin{tabular}[c]{@{}c@{}}11,914\end{tabular}} &53.35& -     & -    & \textbf{99.79} & \textbf{56.77} & \textbf{55.75} \\ 
                                                                              & H2L Hybrid-ViT              & &\textbf{24.33}& 1     & 2    & 99.29 & 51.00 & 50.01 \\ \hline
                                                                              
\multirow{2}{*}{\begin{tabular}[c]{@{}c@{}}CALFW  \\ (Sunglass)\end{tabular}} & D~~~~~~~DeepFaceEMD \cite{hai2022deepface}         & \multirow{2}{*}{\begin{tabular}[c]{@{}c@{}}12,173\end{tabular}} & 73.90 & -     & -    & \textbf{54.95} & 30.66 & 27.74 \\ 
                                                                              & H2L Hybrid-ViT               && \textbf{29.10} & 1     & 6    & 54.00 & \textbf{31.00} & \textbf{27.87} \\ \hline
\multirow{2}{*}{\begin{tabular}[c]{@{}c@{}}AgeDB  \\ (Mask)\end{tabular}}     & D~~~~~~~DeepFaceEMD \cite{hai2022deepface}         & \multirow{2}{*}{\begin{tabular}[c]{@{}c@{}}15,629\end{tabular}} & 72.42 & -     & -    & \textbf{99.84} & \textbf{39.22} & \textbf{33.18} \\ 
                                                                              & H2L Hybrid-ViT              && \textbf{34.44} & 1     & 1    & 99.28 & 33.93 & 26.69 \\ \hline
\multirow{2}{*}{\begin{tabular}[c]{@{}c@{}}AgeDB  \\ (Sunglass)\end{tabular}} & D~~~~~~~DeepFaceEMD \cite{hai2022deepface}         & \multirow{2}{*}{\begin{tabular}[c]{@{}c@{}}16,409\end{tabular}} & 90.40& -     & -    & \textbf{87.06} & 50.04 & 44.27 \\ 
                                                                              & H2L Hybrid-ViT               && \textbf{33.01} & 1     & 2    & 86.75 & \textbf{51.16} & \textbf{44.88} \\ \hline
\end{tabular}
}
\caption{Actual running times and performance for ST2 computation in face identification under occlusion. Compared to DeepFace-EMD (D), the computation of hybrid-ViTs ({H2L}) is significantly faster. For example, for 11,914 query images of the CALFW (mask), H2L runs at least $\textbf{2}$ times faster.   
}
\label{tab:time_run}
\end{table*}

\end{document}

%% file: math_commands.tex
\usepackage{amsmath,amsfonts,bm}

\def\1{\bm{1}}

\DeclareMathAlphabet{\mathsfit}{\encodingdefault}{\sfdefault}{m}{sl}
\SetMathAlphabet{\mathsfit}{bold}{\encodingdefault}{\sfdefault}{bx}{n}

